# Boosted EfficientNet: Detection of Lymph Node Metastases in Breast Cancer Using Convolutional Neural Network

**Jun Wang[1], Qianying Liu[2], Haotian Xie[3], Zhaogang Yang[4], Hefeng Zhou[1]***

[1]Department of Informatics, King's College London, London, WC2R 2LS, UK

[2]College of Management, Shenzhen University, Shenzhen 518060, China

[3]Department of Mathematics, The Ohio State University, Columbus, OH, 43210, Unites States

[4]Department of Radiation Oncology, University of Texas Southwestern Medical Center, Dallas, TX, 75390, United States

**\* Correspondence:**
Hefeng Zhou
H.Rezin.Zhou@gmail.com

**\* Co-Corresponding author:**
Zhaogang Yang
Zhaogang.Yang@UTSouthwestern.edu

**Abstract**

In recent years, advances in the development of whole-slide images have laid a foundation for the utilization of digital images in pathology. With the assistance of computer images analysis that automatically identifies tissue or cell types, they have greatly improved the histopathologic interpretation and diagnosis accuracy. In this paper, the Convolutional Neutral Network (CNN) has been adapted to predict and classify lymph node metastasis in breast cancer. Unlike traditional image cropping methods that are only suitable for large resolution images, we propose a novel data augmentation method named Random Center Cropping (RCC) to facilitate small resolution images. RCC enriches the datasets while retaining the image resolution and the center area of images. In addition, we reduce the downsampling scale of the network to further facilitate small resolution images better. Moreover, Attention and Feature Fusion (FF) mechanisms are employed to improve the semantic information of images. Experiments demonstrate that our methods boost performances of basic CNN architectures. And the best-performed method achieves an accuracy of 97.96% and an AUC of 99.68% on RPCam datasets, respectively.

# 1 Introduction

Even though excellent progress has been made in understanding cancers and blooming the diagnostic and therapeutic methods, breast cancer is the most common malignant cancer diagnosed worldwide, leading to the second cause of cancer-associated death in women[1-3]. Metastatic breast cancers (MBCs), the leading cause of breast cancer death due to their incurable nature, start spreading from the local invasion of surrounding tissues, expand into the lymphatic and blood vessels, and terminate to distant organs[4]. It is estimated that 10% to 50% of patients arise metastases despite diagnosed with regular BC at the beginning[5]. Besides, the rate and site of metastasis possess heterogeneities due to the primary tumor subtype[6]. Thus, accurate diagnosis, prognosis, and treatment for MBCs remain challenging. For BC diagnosis, one of the essential jobs is the staging of BC that counts the recognition of axillary lymph node (ALN) metastases, which is detectable in most node-positive patients using sentinel lymph node (SLN) biopsies[7,8]. Evaluating microscopy images from SLNs are conventional techniques to assess ALNs. However, they require on-site pathologists to investigate samples, which is time-consuming, laborious, and lesser reliable due to a certain degree of subjectivity, particularly in cases that contain small lesions or the lymph nodes are negative for cancer[9].

Consequently, developing digital pathology methods to assist in microscopic diagnosis has evolved significantly during the last decade[10,11]. Advanced scanning technology, cost reduction, quality of spatial images, and magnification have made full digitalization feasible for evaluating histopathologic tissues[12]. Digital pathology has multiple advantages, including remote consultation and sample analysis, thus improving the availability of samples and waiving on-site experts. Still, it requires manual inspection, which brings inconsistent diagnostic decisions caused by individual pathologists that affect the accuracy of diagnosis are unsettled. In addition, hospitals are short of professional equipment and pathologists to support digital pathology. It is reported that presumptive treatment phenomena may exist widely among developing countries due to the lack of well-trained pathologists and professional equipment[13]. Moreover, the majority of the population can barely get access to pathology and laboratory medicine services. Take cancer and cardiovascular disease as examples, only a few and unbalanced communities can get the PLAM treatment[14-16].

To better facilitate digital pathology, reduce the cost of hospitals, and alleviate the problems mentioned before, various analysis methods have been proposed (e.g., deep learning, machine learning, and some specific software) to enhance the accuracy and sensitivity of metastatic cancer detection[17-21]. Convolutional Neural Network (CNN) is the most successful deep learning method in the Computer Vision field due to its robust feature extraction ability. It has been wildly used in diseases diagnosed with microscopy (e.g., Alzheimer's diseases)[22-25]. CNN automatically learns image features from multiple dimensions on a large image dataset, which is applied to identify or classify structures and is therefore applicable in multiple automated image-recognition biomedical areas[26,27]. CNN-based cancer detection was proved as a convenient method to classify tumours from other cells or tissues and has demonstrated satisfactory results[28-31]. EfficientNet is one of the most potent CNN architecture that utilizes the compound scaling method to enlarge the network depth, width, and resolution, obtaining state-of-the-art capacity in various benchmark datasets while requiring lesser computation resources than other models[32]. Hence, the EfficientNet as a suitable model may show significant medical image classification potentials, although there is a big difference between the medical images and traditional images. However, few studies have explored the performance of EfficientNet in medical images, which motivates us to conduct this research.

In this work, we propose three strategies to improve the capability of EfficientNet, including developing a cropping method called Random Center Cropping (RCC) to retain significant features on the center area of images, reducing the downsampling scale of EfficientNet to facilitate the small resolution images of RPCam datasets, and integrating Attention and Feature Fusion mechanisms with EfficientNet to obtain features containing rich semantic information. This work has three main contributions: (1) To our limited knowledge, we are the first study to explore the power of EfficientNet on MBCs classification, and elaborate experiments are conducted to compare the performance of EfficientNet with other state-of-the-art CNN models, which might offer inspirations for researchers who are interested in image-based diagnosis using DL; (2) We propose a novel data augmentation method RCC to facilitate the data enrichment of small resolution datasets; (3) All of our four technological improvements boost the performance of original EfficientNet. The best accuracy and AUC achieve 97.96% and 99.68%, respectively, confirming the applicability of utilizing CNN-based methods for BC diagnosis.

## 2   Related Work

Digital pathology has been widely employed for early cancer detection, classification, and monitoring treatment-response since it can be deployed readily and alleviate the uneven distribution of medical experts to a certain extent while saving their valuable time[33]. The manual process of recognizing MBCs requires high professionalism and many auxiliary materials (e.g., Bone scanning, liver ultrasonography, and chest radiography), and it is time-consuming[34]. In addition, judgments may be affected by some factors, such as fatigue. Due to the unexpected low accuracy of manual-based MBC detection, the arbitration of conflicting double reading opinions in 2005 was put forward[35]. Until now, the challenge of improving diagnostic accuracy is remaining. Therefore, computer-aided diagnosis (CAD) systems were adopted to assist pathologists in interpreting medical images to mitigate problems mentioned before[36].

The traditional machine learning (ML) method plays a crucial role in the MBC classification based on CAD in the early stage. In 1993, Wu et al. used three-layer, feed-forward artificial neural networks to diagnose BC on mammograms, and obtained a ROC value over 95%, which outperformed the average capacity of attending and resident radiologists alone[37]. In 1996, Quinlan applied the decision tree method in BC classification and demonstrated a 94.74% classification accuracy using the C4.5 decision tree with a 10-fold cross-validation[38]. Besides, Hamilton et al.[39] showed a 96% accuracy via the RIAC method, while Ster and Dobnikar[40] gained a 96.8% accuracy via the linear discreet analysis method. Furthermore, Abonyi and Szeifert[41] adopted the supervised fuzzy clustering (SFC) technique and achieved a 95.57% accuracy. However, training and testing datasets in these works are small, leading to low generalization ability.

With the rapid development of computer vision technology, computer hardware, and big data technology, image recognition based on DL has matured. Since AlexNet[42] won the 2012 ImageNet competition, an increasing number of ConvNets have been proposed (e.g., VGG[43], Inception[44], ResNet[45], DenseNet[46]), leading to a significant advance in Computer Vision tasks, including image classification and object detection. Deep Convolutional Neural Networks (DCNNs) models can automatically learn image features, classify images in various fields, and possess higher generalization ability than traditional ML methods, which can distinguish different types of cells, allowing diagnosing other lesions. This technology has also achieved remarkable advances in medical fields[47]. In past decades, many articles have been published relevant to applying the CNN method to cancer detection and diagnosis. For instance, Albayrak et al.[48] developed a CNN-based feature extraction algorithm to detect mitosis in BC histopathological images. In this algorithm, the CNN model was used to extract features to train a support vector machine (SVM) for mitosis detection. Also, DL technology was proved to be useful in lung detection on various image modalities. DCNNs were adopted to predict patients' survival time directly from lung cancer pathological images[49]. Moreover, other groups utilized DL methods to finish medical image classification and achieved good results[50-53].

For BC detection and diagnosis, Agarwal et al.[54] released a CNN method for automated masses detection in digital mammograms, which used transfer learning with three pre-trained models (e.g., VGG16, ResNet50, and InceptionV3). In 2018, Ribli et al. proposed a Faster R-CNN model-based method for the detection and classification of BC masses[55]. The evaluation of their model on the INbreast dataset showed an AUC over 95%. Besides, Shayma'a et al. used AlexNet and GoogleNet to test BC masses on the National Cancer Institute (NCI) and MIAS database[47]. AlexNet performed an accuracy of 97.89% with AUC of 98.32%, and an accuracy of 98.53% with AUC of 98.95% on the National Cancer Institute (NCI) and MIAS database, respectively. In comparison, GoogleNet achieved a 91.58% accuracy with 96.50% AUC and an 88.24% accuracy with 94.65% AUC. Also, Alantari et al. presented a DL method including detection, segmentation, and classification of BC masses from digital X-ray mammograms[56]. They utilized the CNN architecture YOLO and obtained 95.64% accuracy and an AUC of 94.78%[57]. In 2019, Shen et al. came up with a DL method combined with an ensembling strategy to detect and classify BC masses and demonstrated successful results[58]. Besides, Coudray et al. trained a DCNN model based on Google's Inception V3 using full-scale images from Cancer Genome Atlas and further classified images into Adenocarcinoma(LUAD), squamous cell carcinoma (LUSC), or normal tissue automatically[59]. The accuracy of the method was competitive to that of pathologists, with the AUC of 97.00%.

Tan et al. proposed EfficientNet, the state-of-the-art DCNN, that maintains competitive performance while requiring remarkably lesser computation resources in image recognitions[32]. They presented a systematic study to balance the network depth, width, and resolution. Great success could be seen about applying EfficientNet in many benchmark datasets. Academics also explored the capability of EfficientNet in medical imaging classification. Gonçalo Marques et al. utilized EfficientNet to support the diagnosis of COVID-19 and demonstrate a 99.62% accuracy[60]. Miglani V et al. applied EfficientNet to the skin lesion classification and claimed the superiority of EfficientNet over another state-of-the-art model named ResNet50[61].

This work also utilizes EfficientNet as the backbone, which is similar to some aforementioned works, but we focus on the MBC task. In addition, quite different from past works that usually use BC masses datasets with large resolution, our work detects the lymph node metastases in breast cancer and the dataset resolution is small. To our limited knowledge, there is no research to explore the performance of EfficientNet in the detection of lymph node metastases in breast cancer. Therefore, this work aims to examine and improve the capacity of EfficientNet in BC detection.

## 3 Materials and Methods

### 3.1 Rectified Patch Camelyon datasets

A Rectified Patch Camelyon (RPCam) dataset created by deleting duplicate images in Patch Camelyon (PCam) dataset[62,63] is sponsored by the Kaggle Competition. Some images of RPCam datasets are shown in **Figure 1.** The resolution of images in RPCam is 96*96, and the potential pathological features for classifying the cancerous tissues are in the center area with 32*32 size, as shown in the dashed red rectangular of **Figure 1.** RPCam consists of 130908 non-cancer cell images and 89117 cancer cell images.

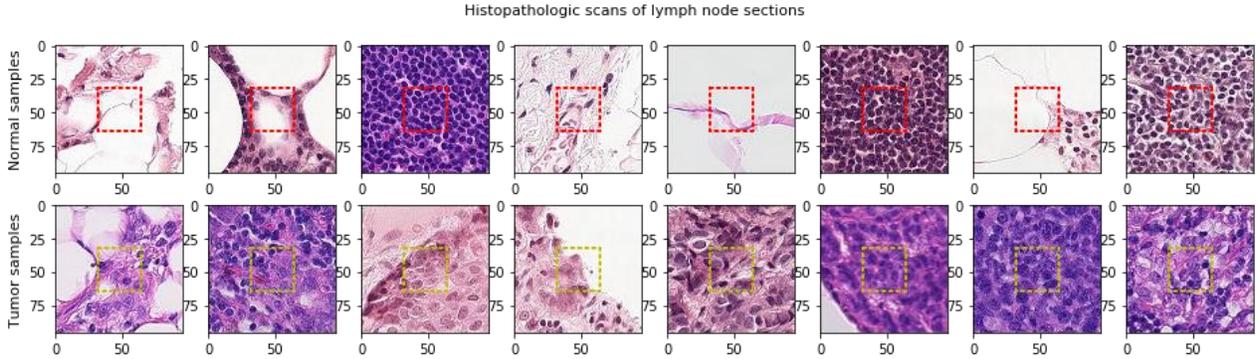

**Figure 1.** Lymph node sections extracted from digital histopathological microscopic scans. Significant features to determine the cancer cells are in the center area (32*32).

### 3.2 Random Center Cropping

The performances of DL models are highly dependent on the scale and quality of training datasets. A large dataset allows researchers to train deeper networks and improves the generalization ability of models, thus enhancing the performance of DL methods. However, establishing large datasets is time-consuming and not economically proficient. To cope with this problem, data augmentation has been proposed to enrich the dataset without introducing new data. Cropping is one of the most commonly used data augmentation methods in Computer Vision tasks and is adopted in our work. However, as mentioned in 3.1, features used for metastasis distinguishments are mainly focused in the central area (32*32) in an image, so traditional cropping methods (Random Cropping and Center Cropping) may lead to the incomplete or lose of these essential areas. Therefore, we propose a cropping method named Random Center Cropping (RCC) to ensure the integrity of the central 32*32 area while selecting peripheral pixels randomly, allowing dataset enrichment. Apart from retaining the significant center areas, RCC maintains more pixels facilitating small resolution images and enabling deeper network architectures. **Figure 2.** illustrates the workflow of the RCC. Specifically, 96*96 images are first enlarged by 8 pixels to 112*112 images via padding 8 pixels around the images, as shown in **Figure 2. (a) to (b)**. Traditional random cropping is then applied to crop these images into 96*96 images.

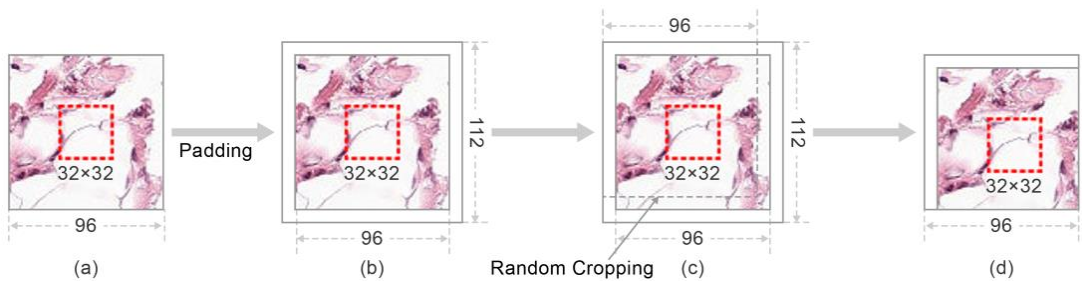

**Figure 2.** The workflow of the Random Center Cropping. Images are first padded eight pixels from four directions (left, right, up, bottom) to create 112*112 resolution. Modified images are then performed random cropping to restore 96*96 resolution. Particular center areas are retained after the cropping process.

### 3.3 Boosted EfficientNet

This section clearly describes our methods to improve the performance of EfficientNet on RPCam datasets. We reduce the downsampling scale to maintain appropriate-level semantics information of features. Besides, Feature Fusion (FF) and Attention mechanisms are embedded in this work, which enhance the feature representation ability and increase the response of vital features. There are eight types of EfficientNet from EfficientNet-B0 to EfficientNet-B7 with an increasing network scale. EfficientNet-B3 is selected as our backbone network due to its superior performances than other architectures according to our experimental results on RPCam datasets.

The architecture of boosted EfficientNet-B3 is shown in **Figure 3.** The main building block is MBConv[64]. Components in red dashed rectangles are different from the original EfficientNet-B3. Images are first sent to some blocks containing multiple convolutional layers to extract image features. Then, these features are weighted by the attention mechanism to improve the response of features contributing to classification. Next, feature fusion mechanism is utilized, enabling features to retain some low-level information. In consequence, images are classified according to those fused features.

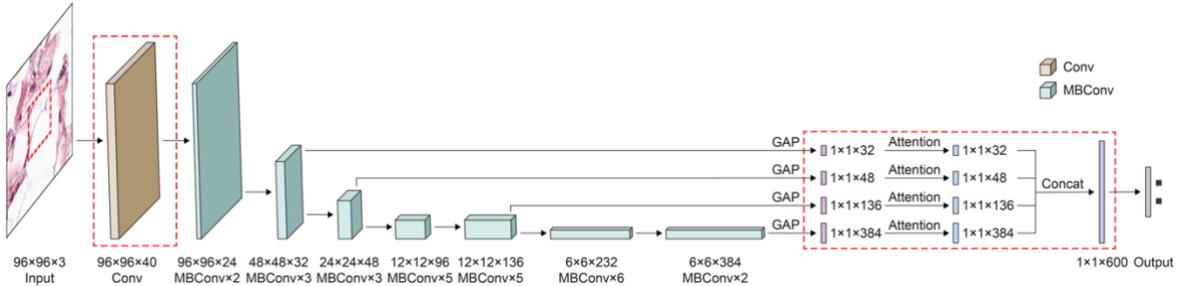

**Figure 3.** The architecture of boosted-EfficientNet-b3. EfficientNet first extracts image features by its convolutional layers. Attention mechanism is then utilized to reweight features, increasing the activation of significant parts. Next, we perform FF on the outputs of several convolutional layers. After that, images are classified based on those fused features. Details of these methods are described in the following sections.

### 3.3.1 Reduce the Downsamping Scale

Although EfficientNet has demonstrated competitive functions in many tasks, we observe that there is a disparity in image resolution between the designed model inputs and RPCam datasets. Most models set their input resolution to 224*224 or lager, which maintains the balance between the performance and time complexity. The depth of the network is also designed for adapting the input size. This setting performs well in most well-known baseline image datasets (e.g., ImageNet[65], PASCAL VOC[66]) as their resolutions usually are more than 1000*1000. However, the resolution of RPCam datasets is 96*96, which is much smaller than the designed model inputs 300*300. After the EfficientNet processing, the size of the final feature will be 32 times smaller than the input (from 96*96 to 3*3). This feature map is likely to be too abstractive and thus losing low-level features, which may defect the performance of EfficientNet.

To mitigate this problem, we adjust the down-sampling multiple in EfficientNet. Our idea is implemented by modifying the stride of the convolution kernel of EfficientNet even though the receptive filed of convolution kernels might be reduced. However, the reduction influence could be slight since the resolution of inputs is small. To select the best-performed downsampling scale, multiple and elaborate experiments are conducted on the downsampling scale {2, 4, 6, 8, 10}, and strategy 16 outperforms other settings. The size of the feature map in best-performed downsampling scale (16) is 6*6, which is one times larger than the original downsampling multiple (32). The change of the downsampling scale from 32 to 16 is implemented by modifying the stride of the first convolution layer from two to one, as shown in the red dashed rectangles on the left half of **Figure 3**.

**3.3.2 Attention Mechanism**

When seeing a picture, the human visual system selectively focuses on a specific part of the picture while ignoring other visible information due to limited visual information processing resources. For example, although the sky information largely covers in the figure, people are able to capture the aeroplane in the image readily (**Figure 4**)[67]. To simulate this process in artificial neural networks, attention mechanism is proposed and has achieved great success in many tasks such as image caption[68,69], image classification[70], and object detection[71,72]. Attention technique can be simply interpreted as a means of increasing the response of the most informative parts and suppressing the activation of others. For instance (**Figure 5**), it can be seen that the response of background is large as most parts of image are background. However, this information usually is useless to the classification, so their response should be suppressed. On the other hand, cancerous tissue is more informative and deserves higher activation, so its response is enhanced after processed by the attention mechanism. As we stated before, the most informative features are in the center area of images on RPCam datasets, making Attention more critical for this work. Hence, this project also adopts the Attention mechanism implemented by a Squeeze-and-Excitation block proposed by Hu et al.[73] Briefly, the essential components are the Squeeze and Excitation. Suppose feature maps $U$ have $C$ channels and the size of the feature in each channel is $H * W$. For Squeeze operation, global average pooling is applied to $U$, enabling features to gain a global receptive field. After Squeeze operation, the size of feature maps $U$ change from $H * W * C$ to $1 * 1 * C$. Results are denoted as $Z$. More precisely, this change is given by

$$\boldsymbol{Z_c = F_{sq}(U_c) = \frac{1}{H \times W} \sum_{i=1}^{W} \sum_{j=1}^{H} U_c(i,j)} \qquad (1)$$

Where $c$ denotes $c^{th}$ channel of $U$, and $F_{sq}$ is the Squeeze function.

Following the Squeeze operation, the Excitation operation is to learn the weight (scalar) of different channels, which is simply implemented by the gating mechanism. Specifically, two fully connected layers are employed to learn the weight of features and activation function sigmoid, and RELU are applied for non-linearity increasing. Excepting the non-linearity, the sigmoid function also certifies the weight falls in the range of [0, 1]. The calculation process of the scalar (weight) is shown in Equation (2).

$$\boldsymbol{S = F_{ex}(Z, W) = \sigma\big(g(Z, W)\big) = \sigma(W_2 \delta(W_1 Z))} \qquad (2)$$

Where $S$ is the result of Excitation operation, $F_{ex}$ is the Excitation function, and $g$ refers to the gating function. $\sigma$ and $\delta$ denote the sigmoid and RELU function, respectively. $W_1$ and $W_2$ are learnable parameters of the two fully connected layers. The final output is calculated by multiplying the scalar S with the original feature maps U.

In our work, the attention mechanism is combined with the feature fusion technique, as shown in **Figure 6**.

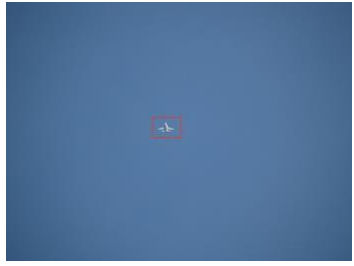

**Figure 4.** Attention in the human visual system. Although the image is mainly covered by the background, people are able to capture the aeroplane in the red rectangular easily.

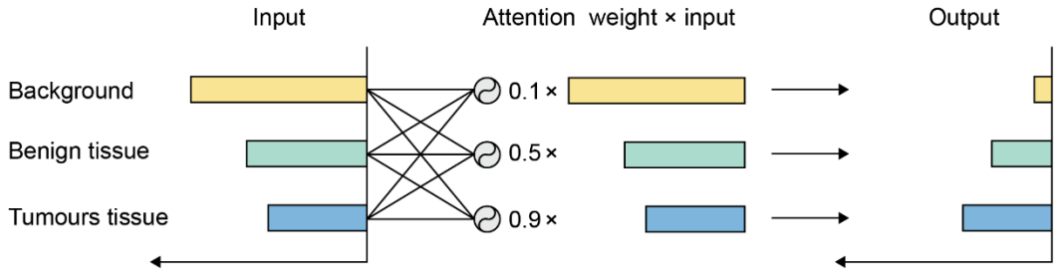

**Figure 5.** An example of the Attention mechanism. The response of input is reweighted by the Attention mechanism.

### 3.3.3 Feature Fusion

High-level features generated by deeper convolutional layers contain rich semantic information, but they usually lose details such as positions and colors that are helpful in the classification. In reverse, low-level features include more detailed information but introducing non-specific noise. FF is a technique that combines low-level and high-level features and has been adopted in many image recognition tasks for performance improvement[74]. Detail information is more consequential in our work since complex textures contours exist in the RPCam images despite their small resolution. Accordingly, we adopt the FF technique to boost classification accuracy.

Four steps are involved during the FF technique (**Figure 6**): 1) During the forward process, we save the outputs (features) of the convolutional layers in the $4^{th}, 7^{th}, 17^{th}$ and $25^{th}$ blocks. (2) After the last convolutional layer extracts features, Attention mechanism is applied to features recorded in step one to value the essential information. (3) Low-level and high-level features are combined using the outputs of step 2 after the Attention mechanism. (4) These fused features are then sent to the following layers to conduct classification.

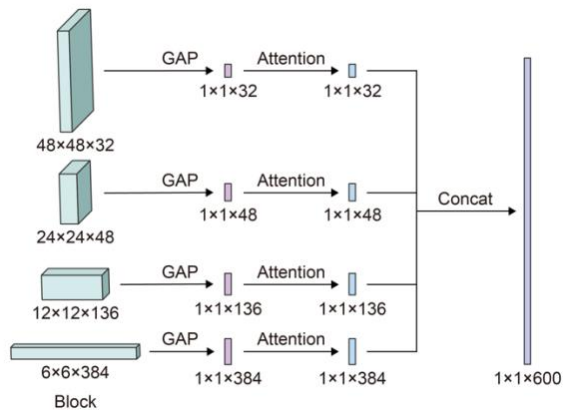

**Figure 6.** The approach to integrate Attention and Feature Fusion mechanisms to EfficientNet-B3.

# 4  Experiments and Results Analysis

This section first introduces the evaluation metrics used for verifying the performance of our methods. Implementation details are then clearly described. Next, we exhibit the capacity of boosted EfficientNet and comparisons among other state-of-the-art models. After that, the influence of each method is investigated via ablation studies. Consequently, elaborate experiments are conducted to explore the effectiveness of the boosted EfficientNet.

## 4.1  Experimental setup

### 4.1.1 Evaluation Metrics

We evaluate our method on the Rectified Patchc Camelyon (RPCam) dataset. Since the testing set is not provided, we split the original training set into a training set and a validation set and utilize the validation set to verify the performance of models. In detail, the capacities of models are evaluated by five indicators, including Area Under the Curve (AUC), accuracy (ACC), sensitivity (SEN), specificity (SPE), and F-measure[75]. AUC considers both Precision and Recall, thus comprehensively reflecting the performance of a model. The value of AUC falls in the range 0.5 and 1, and higher value demonstrates better performances. SEN represents the proportion of all positive examples that are correctly classified and measures the ability of classifiers to recognize positive examples, whereas SPE evaluates the ability of algorithms to recognize negative ones. Like the AUC, F-Measure considers Precision and Recall and is calculated by the weighted average of Precision and Recall. All indicators are calculated based on four fundamental indicators: True Positive (TP), True Negative (TN), False Positive (FP), False Negative (FN). The specific calculation processes are shown in Equation (3) to (6).

$$ACC = \frac{TP+TN}{TP+FP+TN+FN} \qquad (3)$$

$$SEN = \frac{TP}{TP+FN} \qquad (4)$$

$$SPE = \frac{TN}{TN+FP} \qquad (5)$$

$$Fmeasure = \frac{2TP}{2TP+FN+FP} \qquad (6)$$

### 4.1.2 Implementation details

Our method is built on the EfficientNet-B3 model and implemented based on the PyTorch deep learning framework using Python[76]. Four pieces of GTX 2080Ti GPUs are employed to accelerate the training. All models are trained for 30 epochs. The gradient optimizer is Adam. Before being fed into the network, images are normalized by the mean and standard deviation on their RGB-channels. In addition to the RCC, we also employ random horizontal and vertical flipping in the training time to enrich the datasets. During the training, the initial learning rate is 0.003 and decayed by a factor of 10 at the 15th and 23rd epochs. The batch size is set to 256. The parameters of boosted EfficientNet and other comparable models are placed as close as possible to enhance the credibility of the comparison experiment. In detail, the parameter sizes of these three models are increased in turn from the improved EfficientNet, DenseNet121, and ResNet50.

## 4.2  The performance of boosted EfficientNet-B3

Experiments are conducted on the basic EfficientNet and boosted-EfficientNet to evaluate the effectiveness of our methods. Moreover, we compare boosted EfficientNet with another two state-of-the-art CNN models, ResNet50 and DenseNet121[43], to prove its superiority further. The results are shown in **Table 1** and **Figure 7**. It can be seen that basic EfficientNet outperforms boosted-EfficientNet-B3 on the training set both on the ACC and AUC, while a different pattern can be seen on the testing set. The main reason for this different trend is that the basic EfficientNet overfits the training set but boosted-EfficientNet-B3 mitigates overfitting problems since RCC enables the algorithm to crop images randomly, and thus improving the diversity of training images. Although enhancing the performance of a well-performing model is of great difficulty, compared with basic EfficientNet-B3, boosted-EfficientNet-B3 significantly improves the ACC from 97.01% to 97.96% and boosts AUC from 99.24% to 99.68% modestly. Besides, more than 1% increasing can be seen in the SEN, SPE, and F-Measure.

Same patterns of comparison between basic EfficientNet and boosted EfficientNet-B3 can be found when comparing EfficientNet-B3 to other CNN architectures. Notably, ResNet50 and DenseNet121 significantly suffer from the overfitting problem. EfficientNet-B3 obtains better performance than ResNet50 and DenseNet121 for all indicators on testing datasets while using lesser parameters and computation resources (**Figure 7**). All these results confirm the capability of our methods, and we believe these methods can boost other state-of-the-art backbone networks. Therefore, we intend to extend the application scope of these methods in the future. Ablation studies are conducted to illustrate the effectiveness and coupling degree of the four methods, as shown in Section 4.3.

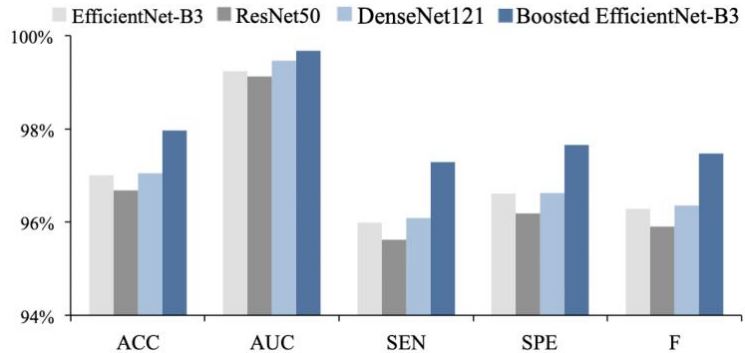

**Figure 7.** Classification results (%) of different methods on the RPCam dataset. Boosted EfficientNet outperforms the other three models in all evaluation metrics.

**Table 1.** Classification results (%) of different methods on the RPCam dataset.

|  | Training | | Test | | | | |
| --- | --- | --- | --- | --- | --- | --- | --- |
|  | ACC | AUC | ACC | AUC | SEN | SPE | F |
| EfficientNet-B3 | 99.61 | 99.99 | 97.01 | 99.24 | 95.99 | 96.61 | 96.29 |
| ResNet50 | **99.85** | **100.00** | 96.68 | 99.13 | 95.62 | 96.18 | 95.90 |
| DenseNet121 | 99.78 | **100.00** | 97.05 | 99.47 | 96.08 | 96.62 | 96.35 |
| **Boosted EfficientNet-B3** | 98.02 | 99.74 | **97.96** | **99.68** | **97.29** | **97.65** | **97.47** |

## 4.3 Ablation studies

In this part, we conduct ablation experiments to illustrate the capacity of our methods, including Random Center Cropping (RCC), Reduce the Downsampling Scale (RDS), Feature Fusion (FF), and Attention. AUC and ACC are utilized as the primary evaluation metrics.

### 4.3.1 The influence of Random Center Cropping

From the first two rows of Table 2, we can observe that the RCC significantly improves performances of algorithms by noticing the AUC is increased from 99.24% to 99.54%, and the ACC is increased from 97.01% to 97.57% because RCC enhances the diversity of training images and mitigates overfitting problem.

### 4.3.2 The influence of Reducing Downsampling Scale

As the first and third rows of Table 2 show, modest improvements of ACC and AUC (0.35% and 0.19%, respectively) can be seen because of the larger feature map. The image resolution of the RPCam dataset is much lower than the designed input of the EfficientNet-B3, resulting in smaller and abstractive features, thus defecting the performance. It is worth noting that the improvement of the RDS is enhanced when being combined with the RCC.

### 4.3.3 The influence of Feature Fusion

Feature fusion (FF) combines low-level and high-level features to boost the performance of models. As shown in Table 2, when adopting only one mechanism, the FF demonstrates the largest AUC and the second-highest ACC increasing among RCC, RDS, and FF, indicating FF's adaptability and effectiveness in EfficientNet. The FF contributes to more remarkable improvement to the model after utilizing RCC and RDS since ACC reaches the highest value, and AUC comes the second among all methods.

### 4.3.4 The influence of Attention Mechanism

It should be emphasized that the Attention mechanism needs to be combined with FF in our work. Utilizing the Attention mechanism to enhance the response of cancerous tissues and suppress the background can further boost the performance. From the 4th, 5th rows of Table 2, it can be seen that the Attention mechanism improves the performance of original architectures both in the ACC and AUC, confirming its effectiveness. Then, we analyze the last four rows. When the first three strategies are employed, adding Attention increases the AUC by 0.02%, but the ACC remains at a 97.96% value. Meanwhile, Attention brings a significant performance improvement comparing with models only utilize RCC and FF since ACC and AUC are increased from 97.59% to 97.85% and from 99.58% to 99.68%, respectively. Although the model using all methods demonstrates the same value of the AUC as the model only utilizing RCC, RDS, and FF, all utilized model shows 0.11% ACC improvements. A possible reason for the minor improvement between these two models is that RDS enlarges the size of the final feature maps, thus maintaining some low-level information to some extent, which is similar to FF and attention mechanism.

**Table 2.** Classification performance comparison of EfficientNet with various strategies on RPCam testing.

|  | **RCC** | **RDS** | **FF** | **Attention** | **ACC (%)** | **AUC (%)** |
|---|---|---|---|---|---|---|
| EfficientNet |  |  |  |  | 97.01 | 99.24 |
|  | √ |  |  |  | 97.57 | 99.54 |
|  |  | √ |  |  | 97.36 | 99.43 |
|  |  |  | √ |  | 97.55 | 99.57 |
|  |  |  | √ | √ | 97.63 | 99.63 |
|  | √ | √ |  |  | 97.73 | 99.62 |
|  | √ | √ | √ |  | **97.96** | 99.66 |
|  | √ | √ | √ | √ | **97.96** | **99.68** |
|  | √ |  | √ |  | 97.59 | 99.58 |
|  | √ |  | √ | √ | 97.85 | **99.68** |

## 5 Conclusion

The purpose of this project is to facilitate the development of digital diagnosis in MBCs and explore the applicability of a novel CNN architecture EfficientNet on MBC. In this paper, we propose a boosted EfficientNet CNN architecture to automatically diagnose the presence of cancer cells in the pathological tissue of breast cancers. We develop a data augmentation method RCC to retain the most informative parts of images and maintain original image resolution. Experiments demonstrate that this method significantly improves the performance of EfficentNet-B3. In addition, we propose to reduce the downsampling scale of basic EfficientNet by adjusting the architecture of EfficientNet-B3 to facilitate small resolution training images better. Moreover, two mechanisms are employed to enrich the semantic information of features. As shown in the ablation studies, both of these methods boost the basic EfficientNet-B3, and more remarkable improvements can be obtained by combining some of them. Boosted-EfficientNet-B3 is also compared with another two state-of-the-art CNN architectures, ResNet50 and DenseNet121, and shows superior performance. We believe that our methods can be utilized in other models and lead to improved performance on other diseases diagnosis and will explore this in the future. In summary, our boosted EfficientNet-B3 achieves an accuracy of 97.96% and an AUC value of 99.68%, respectively, and hence may provide a reliable, efficient, and economical alternative for medical institutions in relevant areas.

**Data Availability**

All data generated or analyzed during this study are included in this published article and its supplementary information files.

**Conflicts of Interest**

The authors declare that they have no competing interests.

**Authors' Contributions**

J Wang and HF Zhou conceived and designed the experiments. J Wang, HF Zhou, QY Liu, HT Xie, ZG Yang analyzed and extracted data. HF Zhou and QY Liu constructed figures. HF Zhou, QY

Liu participated in table construction. All authors participated in the writing, reading, and revising of the manuscript and approved the final version of the manuscript.